\documentclass[11pt]{article}
\usepackage{amssymb}
\usepackage[preprint]{acl}
\usepackage{tikz}
\usepackage[table]{xcolor}
\usetikzlibrary{shapes.geometric, arrows.meta, positioning, fit, backgrounds, calc}
\usepackage{times}
\usepackage{latexsym}
\usepackage{amsmath}
\usepackage{multirow}
\usepackage{booktabs}
\usepackage[T1]{fontenc}

\usepackage[utf8]{inputenc}

\usepackage{microtype}

\usepackage{inconsolata}

\usepackage{graphicx}
\usepackage{eso-pic}
\title{Sampling Reveals Style: Unsupervised, Training-Free Discovery of Prompt-Conditional Stylistic Axes in LLM Activations}

\author{Ajit Mallavarapu \\
   Cornell Tech / New York, NY \\
   \texttt{am3574@cornell.edu} \\\And
   Ziwei Gu \\
   Harvard University / Cambridge, MA \\
   \texttt{ziweigu@g.harvard.edu} \\}

\begin{document}
\setcounter{page}{639}
\AddToShipoutPictureBG*{\setlength{\unitlength}{1mm}\footnotesize
  \put(0,13){\parbox[t]{\paperwidth}{\centering
    \emph{Proceedings of the 9th BlackboxNLP Workshop: Analyzing and Interpreting Neural Networks for NLP}, pages 639\textendash 654\\
    October 29, 2026 \textcopyright 2026 Association for Computational Linguistics}}}
\maketitle 

\begin{abstract}
Large language models (LLMs) encode rich stylistic structure in their hidden activations, but discovering which stylistic dimensions are salient for a given prompt typically requires supervised contrastive data. We present a training-free, prompt-conditional alternative: we repeatedly sample completions of a single prompt at elevated temperature, apply Principal Component Analysis (PCA) to the pooled hidden activations, and label the resulting axes automatically from the pole generations. We validate the discovered axes against 245 human-elicited stylistic annotations in a two-phase study. On our strongest model (Qwen3.5-4B), the top two axes match spontaneously requested human dimensions with 72.8\% precision and 43.6\% macro-recall, and 75.6\% of validity ratings judge the axes' polar generations accurate to their labels, with 90.9\% adjacent inter-annotator agreement. Discoverability is strongly model-dependent: both Qwen models and Llama-3.2-3B expose human-salient axes, while DeepSeek-7B-Chat drops to 35.3\% precision, its leading components dominated by structural rather than stylistic variance. Simple PCA over a model's own decoding variance is thus an effective, low-cost probe of stylistic structure in LLM representations, one that also exposes sharp cross-model differences in how that structure is organized.
\end{abstract}

\section{Introduction}

Large language models (LLMs) encode a vast array of human knowledge within their internal representations. As these models scale, understanding the conceptual structures they naturally form becomes increasingly critical for interpretability and safety \citep{zou2025representationengineeringtopdownapproach}. However, discovering these concepts without relying on manual annotations or predefined taxonomies remains a formidable challenge. The internal states of state-of-the-art models are notoriously complex, making it difficult to isolate distinct, human-interpretable concepts from dense, high-dimensional spaces \citep{cunningham2023sparse,templeton2024scaling}.

In this work, we ask whether the stylistic dimensions most salient for a given prompt can be discovered directly from a model's activations, with no supervision at any stage. Our pipeline repeatedly samples completions of a single prompt at elevated temperature, applies Principal Component Analysis (PCA) to the pooled hidden activations, and labels the surviving components with an LLM judge. The result is a training-free, prompt-conditional method that surfaces the few stylistic axes dominating the model's own decoding variance for that instruction.

To rigorously evaluate the semantic coherence and interpretability of the discovered axes, we conducted a two-phase human evaluation. In the first phase, annotators were asked to spontaneously envision ideal stylistic controls for a given prompt, and our unsupervised pipeline recovered those requested dimensions with strong agreement on our strongest model. In the second phase, human judges evaluated text generated at the geometric extremes of the discovered axes and found that the pole generations closely embodied the automatically generated labels, with 75.6\% of ratings marking them as accurate or highly accurate and 90.9\% adjacent inter-annotator agreement. Together, these results suggest that the discovered axes do not merely capture statistical regularities, but correspond to representations that are semantically meaningful and legible to human annotators.

Crucially, applying our framework across various architectures revealed a surprising representational phenomenon, which we term \textbf{Structural Entanglement}. Specifically, we observed a substantial latent entanglement in the DeepSeek model we evaluate. In this model, stylistic variance exhibits strong, inextricable coupling with structural and syntactic variance in the leading components of the representation space. This entanglement challenges the assumption that stylistic and task variance are linearly separable, highlighting differences in how distinct training regimes or architectures organize their internal representations.

\textbf{Our main contributions are threefold:}
\begin{enumerate}
    \item A training-free, prompt-conditional method for discovering stylistic axes from LLM activations via PCA and automatic labeling.
    \item A two-phase human evaluation showing that the discovered axes align with spontaneous human intent and are judged semantically valid.
    \item An empirical characterization of model-specific sensitivity, including a failure case in \texttt{DeepSeek-7B-Chat} where structural variance overwhelms stylistic variance.
\end{enumerate}

\section{Methodology}

\begin{figure*}[t]
    \centering
    \resizebox{\textwidth}{!}{%
    \begin{tikzpicture}[
        >=Stealth,
        node distance=1.5cm and 2cm,
        basebox/.style={draw, rounded corners, align=center, fill=white, thick, minimum height=1.2cm, minimum width=2.5cm, font=\sffamily\small},
        cloudbox/.style={draw, rounded corners, align=center, fill=gray!10, thick, minimum height=1.2cm, minimum width=2.8cm, font=\sffamily\small},
        modelblock/.style={draw, rectangle, rounded corners, align=center, fill=blue!10, thick, minimum height=1.4cm, minimum width=2.8cm, font=\sffamily\small},
        mathblock/.style={draw, rectangle, align=center, fill=green!10, thick, minimum height=1.2cm, minimum width=3.2cm, font=\sffamily\small},
        judgeblock/.style={draw, rectangle, rounded corners, align=center, fill=orange!10, thick, minimum height=1.2cm, minimum width=2.8cm, font=\sffamily\small},
        grouphighlight/.style={draw=black!50, dashed, rounded corners, inner sep=12pt}
    ]

    \node[basebox] (prompt) {$P$ \\ Base Prompt};
    
    \node[cloudbox, right=2.5cm of prompt, yshift=0.25cm] (v2) {};
    \node[cloudbox, right=2.5cm of prompt, yshift=0.125cm] (v1) {};
    \node[cloudbox, right=2.5cm of prompt] (cloud) {$C = \{p_1, \dots, p_N\}$ \\ Variation Cloud \\ \scriptsize ($N=30$)};
    
    \draw[->, thick] (prompt) -- node[above, font=\sffamily\scriptsize] {Stochastic Generation} node[below, font=\sffamily\scriptsize] {(Auto-Mutation)} (cloud.west);

    \node[modelblock, right=1.5cm of cloud] (target) {Target Model \\ \scriptsize (e.g., Qwen3.5-4B)};
    \node[basebox, below=0.8cm of target] (layer) {Penultimate Layer \\ Activations};
    \node[mathblock, right=1.5cm of layer] (matrix) {Matrix $A \in \mathbb{R}^{N \times d}$ \\ \scriptsize Sequence-Avg Pooling};
    
    \draw[->, thick] (cloud) -- (target);
    \draw[->, thick] (target) -- (layer);
    \draw[->, thick] (layer) -- (matrix);

    \node[mathblock, right=2cm of matrix] (pca) {PCA Decomposition \\ \scriptsize Retain $PC_1, PC_2$};
    
    \draw[->, thick] (matrix) -- (pca);

    \node[judgeblock, right=2cm of pca] (judge) {Llama-3 Judge \\ \scriptsize Auto-Labeling};
    \node[basebox, fill=yellow!10, right=1.5cm of judge] (slider) {\textbf{Interpretable Axis} \\ \scriptsize $[-]$ Sarcasm $[+]$};
    
    \draw[->, thick] (pca) -- node[above, font=\sffamily\scriptsize] {De-noised Axes} (judge);
    \draw[->, thick] (judge) -- node[above, font=\sffamily\scriptsize] {Polarity} node[below, font=\sffamily\scriptsize] {Check} (slider);

    \begin{scope}[on background layer]
        \node[grouphighlight, fill=blue!5, fit=(target)(layer)(matrix), label={[font=\sffamily\bfseries, text=blue!80!black]above:Activation Harvesting}] {};
        \node[grouphighlight, fill=green!5, fit=(pca), label={[font=\sffamily\bfseries, text=green!60!black]above:Latent Component Selection}] {};
        \node[grouphighlight, fill=orange!5, fit=(judge)(slider), label={[font=\sffamily\bfseries, text=orange!80!black]above:Semantic Grounding}] {};
    \end{scope}

    \end{tikzpicture}%
    }
    \caption{The \textbf{Latent Engine} architecture for Unsupervised Concept Discovery. The pipeline isolates stylistic variance through stochastic manifold generation into a variation cloud ($N=30$), extracts representations from the penultimate hidden layer, mathematically isolates the primary axes of variance via PCA by extracting the top two principal components ($PC_1, PC_2$), and autonomously labels these orthogonal components using a lightweight linguistic judge.}
    \label{fig:latent_engine}
\end{figure*}
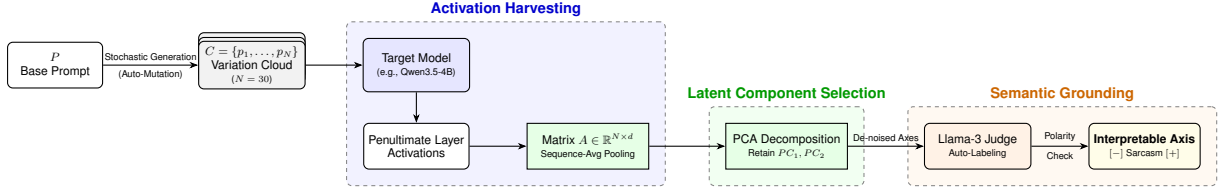

This section details our unsupervised pipeline for discovering and validating latent stylistic axes within Large Language Models (LLMs). Our approach, the \textbf{Latent Engine} (illustrated in Figure \ref{fig:latent_engine}), moves beyond supervised contrastive pairs \citep{rimsky2024steering,konen2024style} by leveraging internal model geometry to surface human-salient concepts.

\subsection{Stochastic Manifold Generation (Variation Clouds)}
To discover latent dimensions without supervised contrastive pairs or explicit stylistic prompting, we isolate the natural geometric variations of a model's activations during stochastic decoding. For a given base prompt $x$, the system generates a ``variation cloud'' $C = \{p_1, p_2, \dots, p_N\}$ where $N=30$.

\begin{itemize}
    \item \textbf{Autoregressive Stochasticity:} Unlike methods that rely on explicit stylistic mutators, our framework generates variations by sampling $N$ independent trajectories from the base model using a single static input prompt.
    \item \textbf{Sampling Parameters:} To maximize stylistic exploration without degrading token-level coherence, we utilize elevated temperature settings ($T=0.9$, $p=0.95$). This ensures that the latent axes emerge naturally from the model's own stochastic decoding distribution, capturing the model's native stylistic manifold for that specific instruction.
\end{itemize}

\subsection{Activation Harvesting \& PCA Engine}
We extract the internal representations of the target model as it processes the variation cloud $C$.
\begin{itemize}
    \item \textbf{Target Layers:} Activations are harvested dynamically from the penultimate hidden layer to maintain consistent relative representational depth across varying architectures, thereby avoiding the structural noise prevalent in earlier layers and the final vocabulary-projection biases of the ultimate layer.
    \item \textbf{Activation Harvesting Window:} For each variation $p_i \in C$, we concatenate the static base prompt $x$ with the stochastically generated response tokens. We then execute a forward pass to extract the hidden states specifically across the generated sequence window (from the first generated response token to the final token prior to the end-of-sequence identifier). This guarantees that our extraction framework isolates downstream token choices and stylistic variance rather than deterministic prompt representations.
    \item \textbf{Sequence Pooling:} We apply sequence-average pooling over the length $L_i$ of the generated response token hidden states for variant $i$. This compresses the dynamic token sequence matrix into a single sequence-averaged activation vector $\mathbf{h}_i \in \mathbb{R}^d$, where $d$ is the hidden dimension of the model. Stacking these vectors across all generations yields the target variation matrix $\mathbf{A} \in \mathbb{R}^{N \times d}$.
    \item \textbf{Latent Extraction:} We perform Principal Component Analysis (PCA) on the target variation matrix $\mathbf{A}$ to identify the orthogonal axes capturing the greatest variance within the style manifold.
\end{itemize}

\subsection{Latent Component Isolation}
Following the PCA decomposition of the variation manifold, the pipeline isolates the top two principal components ($PC_1, PC_2$). By definition, standard PCA yields strictly orthogonal singular vectors ordered by their explained variance. By retaining these primary components, the Latent Engine successfully isolates the most dominant, linearly independent axes of variance that emerged naturally during the stochastic decoding process. These primary vectors capture the core geometric directions of the model's native generation space, surfacing highly targeted latent vectors for subsequent semantic evaluation and stylistic manipulation.

\subsection{Auto-Labeling and Polarity Alignment}
To make the discovered axes interpretable, the pipeline synthesizes ``pole generations'' by intervening on the model's hidden states along each normalized principal component $\hat{\mathbf{v}}_k$ during a new forward pass. Specifically, we apply an additive activation shift
$$
\mathbf{h}' = \mathbf{h} + \alpha \hat{\mathbf{v}}_k,
$$
where $\alpha$ is a dimensionless scalar intervention coefficient controlling the magnitude and sign of the shift. We use $\alpha = \pm 0.6$ to generate the two opposing poles of each discovered axis. The magnitude $\lvert\alpha\rvert=0.6$ was selected empirically: intervention sweeps indicated that generations remained coherent within an interactive steering envelope of approximately $\lvert\alpha\rvert\leq0.75$, while larger magnitudes could produce repetitive or syntactically degraded outputs. We therefore use $\lvert\alpha\rvert=0.6$ as a mid-to-high intervention magnitude that produces differentiated pole generations while remaining within the observed coherence range.

These synthesized extremes are then evaluated in an autonomous grounding step:
\begin{itemize}
\item \textbf{Linguistic Judge:} A quantized Meta-Llama-3-8B-Instruct model examines generations at the extremes of each PCA axis and generates a single, concise label (e.g., `Formality,'' `Sarcasm''). Crucially, this labeling process was conducted completely blind; the judge was provided exclusively with the raw text strings and was stripped of any metadata identifying the target model's family or architecture.
\item \textbf{Dynamic Polarity Alignment:} Because PCA axes possess arbitrary sign orientations, we utilize the same LLM judge to establish intuitive polarity. The judge evaluates the extreme generations to classify which text more strongly embodies the newly discovered label. The axis is then automatically oriented so that the positive direction ($+X$) corresponds to an increase in the labeled attribute.
\end{itemize}

Because PCA component signs are arbitrary, the initial $+\hat{\mathbf{v}}_k$ and $-\hat{\mathbf{v}}_k$ directions should not be interpreted as semantically positive or negative a priori; polarity is assigned only after the semantic labeling step.

\subsection{Target Architectures}
To evaluate the universality of this latent organization, we port the pipeline across four distinct model families: Qwen3.5-4B \citep{qwen35blog}, Qwen2.5-3B \citep{qwen2025qwen25technicalreport}, Llama 3.2 \citep{grattafiori2024llama3herdmodels}, and DeepSeek \citep{iai2024deepseekllmscalingopensource}. This selection allows us to measure \textbf{Structural Entanglement}—the degree to which stylistic variance is natively separable from strict structural and formatting priors across different alignment regimes.
\subsection{Grounded Geometric Semantic Evaluation}
Finally, we validate the discovered axes through a two-phase human evaluation protocol:
\begin{enumerate}
    \item \textbf{Phase 1 (Spontaneous Recall):} Annotators list the ``magic sliders'' they desire for a given prompt, testing if the system's discovered axes match spontaneous human intent. To compute precision and recall against these unconstrained free-text responses, we embed the labels utilizing \textbf{all-mpnet-base-v2} \citep{song2020mpnet} and apply a semantic matching threshold of $\tau = 0.65$ to reliably capture stylistic synonyms without introducing false positives.

    To mathematically account for this linguistic variance, we anchor our baseline semantic alignment threshold at $\tau = 0.65$. While a naive evaluation framework might demand near-perfect embedding similarity ($\tau \geq 0.85$) between the model's auto-generated label and the human target concept, empirical observation of the latent manifolds reveals a persistent \textit{human-machine lexical gap}. Models frequently isolate stylistic control vectors that are geometrically coherent but exhibit slight lexical drift from typical human responses. For instance, an unsupervised axis that a human annotator classifies as \textit{Sarcasm} is frequently labeled by the autonomous pipeline as \textit{Snarkiness} or \textit{Cynicism} ($\text{cos}(\theta) \approx 0.68$). Similarly, vectors capturing \textit{Pragmatism} often surface under the adjacent label of \textit{Directness}. Demanding exact token matches heavily penalizes the system for identifying valid, contiguous semantic manifolds that simply utilize adjacent vocabulary blocks. 
    
    We anchor our baseline semantic alignment threshold at $\tau = 0.65$ as a mid-range setting that balances semantic flexibility with matching specificity. The sensitivity sweep in Table \ref{tab:tau_sensitivity} in Appendix \ref{sec:appendix_threshold} shows a smooth degradation in precision as $\tau$ increases, rather than a distinct plateau or elbow. Under this baseline configuration, Precision is evaluated against the $D = 40$ auto-generated dimensions ($20\text{ prompts} \times 2\text{ components}$), while Macro-Recall is computed at the annotator--prompt level: for each annotator--prompt instance $i$, we measure the fraction of that instance's $n_i$ human-elicited stylistic requests that are recovered by the system, and then average these per-instance recall values. Across the study, annotators provided an average of 3.9 distinct stylistic requests per prompt, yielding 245 unique expressions in aggregate. We therefore use $\tau=0.65$ as a fixed, mid-range evaluation setting rather than claiming it as an empirically optimal threshold.    
    
    \item \textbf{Phase 2 (Polar Validity):} Annotators rate the accuracy of the system's auto-generated labels against actual polar text generations ($-X$ vs $+X$) on a 1--5 Likert scale. We evaluate annotator consensus using Mean Exact Agreement and Mean Adjacent Agreement (where consensus is achieved if annotator ratings fall within $\pm 1$ point of each other on the Likert scale). This effectively bypasses the ordinal skew paradox (the paradox of high agreement) common in traditional reliability metrics.

\end{enumerate}

\section{Results}
\begin{table*}[htbp]
\centering
\small
\begin{tabular}{lp{3.5cm}cp{5.5cm}}
\toprule
\textbf{Evaluation Phase} & \textbf{Metric} & \textbf{Score} & \textbf{Interpretation} \\
\midrule
\multirow{3}{*}{\shortstack[l]{\textbf{Phase 1: Spontaneous Recall}\\\textbf{(Unconstrained)}}} 
& Human Semantic IAA & 47.2\% & Baseline semantic consensus among human annotators. \\
& \textbf{System Precision (Qwen3.5-4B)} & \textbf{72.8\%} & When the system extracts a concept, 72.8\% of the time it matches a human-requested trait at $\tau = 0.65$ to a human-desired trait. \\
& System Macro-Recall & 43.6\% & On average, the system recovers 43.6\% of the stylistic dimensions requested within each annotator–prompt instance. \\
\midrule
\multirow{5}{*}{\shortstack[l]{\textbf{Phase 2: Grounded Validity}\\\textbf{(Constrained)}}} 
& Global Median Score & 4.0 / 5.0 & The typical system-discovered axis is rated ``Mostly Accurate.'' (IQR: 1.0) \\
& \textbf{Top-2 Box Accuracy} & \textbf{75.6\%} & >75\% of all extreme polar generations were rated highly accurate to their auto-generated label. \\
& Statistical Significance & $p < 0.001$ & Wilcoxon Signed-Rank Test confirms axes strictly outperform a neutral baseline. \\
& Mean Exact Agreement & 57.5\% & Absolute majority of annotators selected the exact same 1-5 rating cell. \\
& \textbf{Adjacent Agreement} & \textbf{90.9\%} & 9 out of 10 annotator ratings land within $\pm1$ point, establishing near-universal human consensus. \\
\bottomrule
\end{tabular}
\caption{Human Grounding and Geometric Validity of Unsupervised Latent Axes. Precision and Recall metrics for Qwen3.5-4B represent the robust macro-average across 9 independent stochastic seeds ($N=30$ prompt variations per seed). Meta-Llama-3-8B-Instruct was utilized as the semantic labeler across all runs.}
\label{tab:main_results}
\end{table*}

\subsection{Human Validation and Grounded Geometry}
\label{sec:aggregate_human_validation}

Our evaluation isolates the effectiveness of the unsupervised latent discovery pipeline by testing against spontaneous human desire (Phase 1) and grounded geometric validity (Phase 2). Before examining model-specific differences (detailed in Section 3.4), we first report the aggregate baseline performance across our primary evaluation architecture, summarized in Table \ref{tab:main_results}.

In Phase 1 (Unconstrained Generation), annotators evaluated whether the autonomously discovered latent axes matched their spontaneous conceptualization of stylistic control. Human annotators demonstrated a semantic Inter-Annotator Agreement (IAA) of 47.2\%, establishing a stable ground-truth baseline for human vocabulary alignment. Because our variation clouds are generated using stochastic decoding, all reported system metrics represent the macro-average across 9 independent experimental seeds to ensure robust reproducibility. Against this baseline, our top-performing architecture (\texttt{Qwen3.5-4B}) achieved a \textbf{72.8\% mean Precision}, indicating that nearly three-quarters of the automatically discovered latent axes matched a human-requested trait at $\tau = 0.65$. Additionally, we recorded a Phase 1 \textbf{mean Macro Recall of 43.6\%}. This is a high recovery rate given that the theoretical maximum recall for a Top-2 axis extraction is approximately 51.7\%\footnote{The theoretical maximum recall is constrained by the structural limits of the top-$k$ evaluation setting. In our methodology, the unsupervised pipeline was restricted to surfacing only the top two most salient principal components ($k=2$) per prompt. For each annotator--prompt instance $i$ with $n_i$ human-elicited dimensions, the maximum recoverable fraction is $\min(2,n_i)/n_i$. Thus, the theoretical ceiling is the mean of $\min(2,n_i)/n_i$ across annotator--prompt instances, rather than exactly $2/3.9$. Given the observed distribution of $n_i$, this ceiling is approximately 51.7\%. Our achieved recall of 43.6\% therefore captures approximately 84\% of the available theoretical capacity ($43.6/51.7$). The 245 unique expressions describe the pooled diversity of human responses; recall is computed at the annotator--prompt level.}

In Phase 2 (Polar Validity), annotators judged the semantic validity of the generated extremes on a 1–5 Likert scale. To accurately match auto-generated axis labels with human intents as ``stylistic synonyms'' prior to evaluation, we applied an embedding cosine-similarity threshold of $\tau = 0.65$. Under this threshold constraint, the aggregate results demonstrate strong alignment: the system achieved a \textbf{Global Median Score of 4.0 out of 5.0} ($IQR = 1.0$, $Q_{25} = 4.0$, $Q_{75} = 5.0$). Furthermore, the pipeline recorded a \textbf{Top-2 Box Accuracy of 75.6\%}, meaning more than three-quarters of all human ratings judged the auto-discovered axes as either ``Mostly Accurate'' (4) or ``Highly Accurate'' (5). 

For a qualitative showcase of the auto-discovered stylistic axes, including unedited generation snippets from the geometric extremes ($-X, +X$) of the latent manifold, please refer to Appendix \ref{sec:appendix_qualitative}. 

This positive alignment is statistically strong (Wilcoxon Signed-Rank Test comparing the ordinal ratings against a neutral null-hypothesis baseline of 3.0, $p=1.493\times10^{-42}$) (see Figure \ref{fig:violin_density} in Appendix \ref{sec:appendix_eval}), indicating a very low probability that these dimension alignments represent random noise. Finally, the system achieved a Mean Exact Agreement of 57.5\% and an exceptional \textbf{90.9\% Adjacent Agreement}\footnote{\textbf{Metric Definitions:} \textit{Adjacent Agreement} is an inter-rater reliability metric for ordinal scales that considers scores within one point (e.g., a 4 and a 5) as consensus, thereby preventing penalization for minor subjective differences in intensity judgments.}. This confirms that the PCA-derived axes track the targeted semantic properties—establishing a strong, statistically significant baseline of human agreement before introducing model-specific variations.

\begin{figure}
    \centering
    \includegraphics[width=1\linewidth]{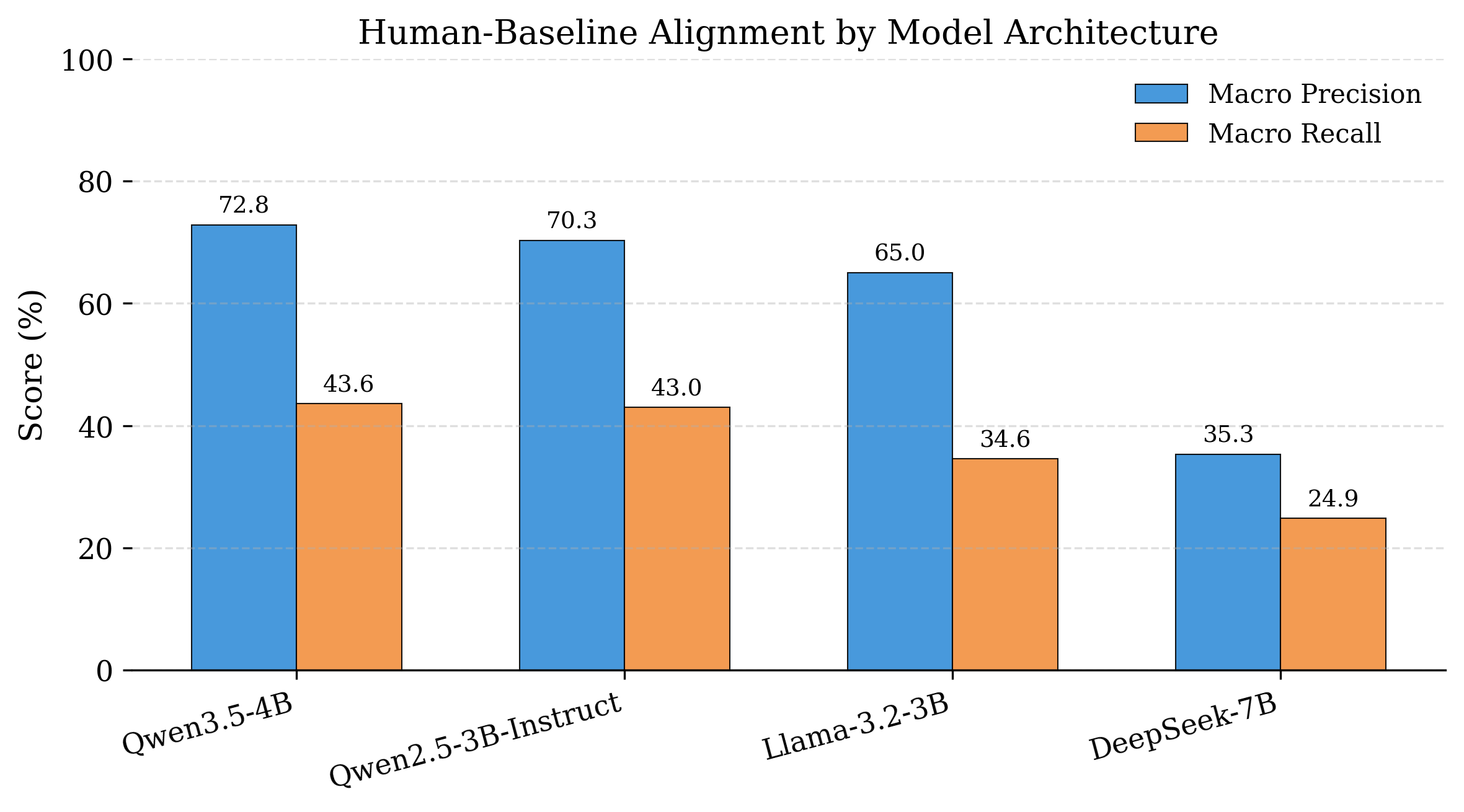}
    \caption{\textbf{Human-Baseline Alignment Across Architectures.} Macro Precision and Recall scores for PCA-discovered stylistic axes evaluated against the Phase 1 human spontaneous recall baseline. While Qwen and Llama architectures successfully expose human-salient latent controls, DeepSeek-7B exhibits a severe interpretability collapse, yielding only 35.3\% precision.}
    \label{fig:human_baseline_alignment}
\end{figure}
\subsection{Stochastic Stability of Discovered Axes}
A key concern in unsupervised representation engineering is whether the discovered latent geometries are stable mechanisms or stochastic artifacts of the generation process. We performed repeated extractions across \textbf{9 experimental seeds} for the variation clouds, yielding a total of 180 prompt–seed extraction runs. Each run produced two principal components (PC1 and PC2), for a total of \textbf{360 individual candidate axes}. Crucially, these variation clouds were generated strictly through elevated temperature sampling ($T$=0.90) from the static base prompt, free of any external mutators. Across these conditions, the \textbf{Latent Engine} demonstrated high robustness, achieving an \textbf{82.5\% Top-2 coverage rediscovery rate}. Specifically, stability across these runs was evaluated by measuring whether the auto-labeled stylistic dimensions—matched via our standard $\tau = 0.65$ semantic similarity threshold—reappeared within the Top-2 principal components ($PC_1, PC_2$) for each prompt. This 82.5\% rediscovery rate represents the proportion of successfully matched concept instances out of the total cross-seed evaluation denominator (360 candidate axes across 20 prompts $\times$ 9 seeds). This indicates that the primary stylistic dimensions in the manifold are structurally entrenched in the model's internal representations (specifically, within the targeted penultimate hidden layer) rather than transient noise.

\subsection{Text-Only Baseline and Surface Confound Control}
To rigorously test whether internal hidden activations contribute unique value beyond surface-level text representation, we implemented a text-only baseline across our experimental framework. Bypassing penultimate layer activations, we applied Principal Component Analysis directly to the embedding matrix of the raw 30-text variation clouds using the \texttt{all-mpnet-base-v2} encoder. Autonomous labeling by the Llama-3 judge revealed that raw text PCA predominantly isolates low-level surface confounds, such as response length and vocabulary overlap, rather than abstract stylistic dimensions.

Quantitatively, across our evaluation cohort of $D=40$ dimensions evaluated against the pooled Phase 1 human baseline from our 8 annotators (who averaged 3.9 stylistic requests per prompt out of a total pool of 245 expressions), the text-only baseline achieved a Macro Precision of 38.4\% and a Macro Recall of 21.2\%. This represents a sharp degradation compared to our activation-based Latent Engine on Qwen3.5-4B, which attained 72.8\% precision and 43.6\% macro-recall. Qualitative inspection confirmed that while text-embeddings entangle stylistic choices with structural heuristics, the model's deep hidden states cleanly organize the manifold along orthogonal stylistic axes like Formality and Sarcasm. This empirical divergence confirms that the Latent Engine's reliance on internal model geometry captures genuine semantic abstractions unachievable through surface-level string clustering.

\subsection{Cross-Architecture Discoverability and the DeepSeek Anomaly}
Applying the Latent Engine across our target architectures revealed highly model-dependent discoverability. The Qwen models demonstrated superior human-baseline alignment, maintaining a precision of approximately 70\%. \texttt{Meta-Llama-3.2-3B} showed moderate but reliable extraction capabilities.

\begin{figure}
    \centering
    \includegraphics[width=1\linewidth]{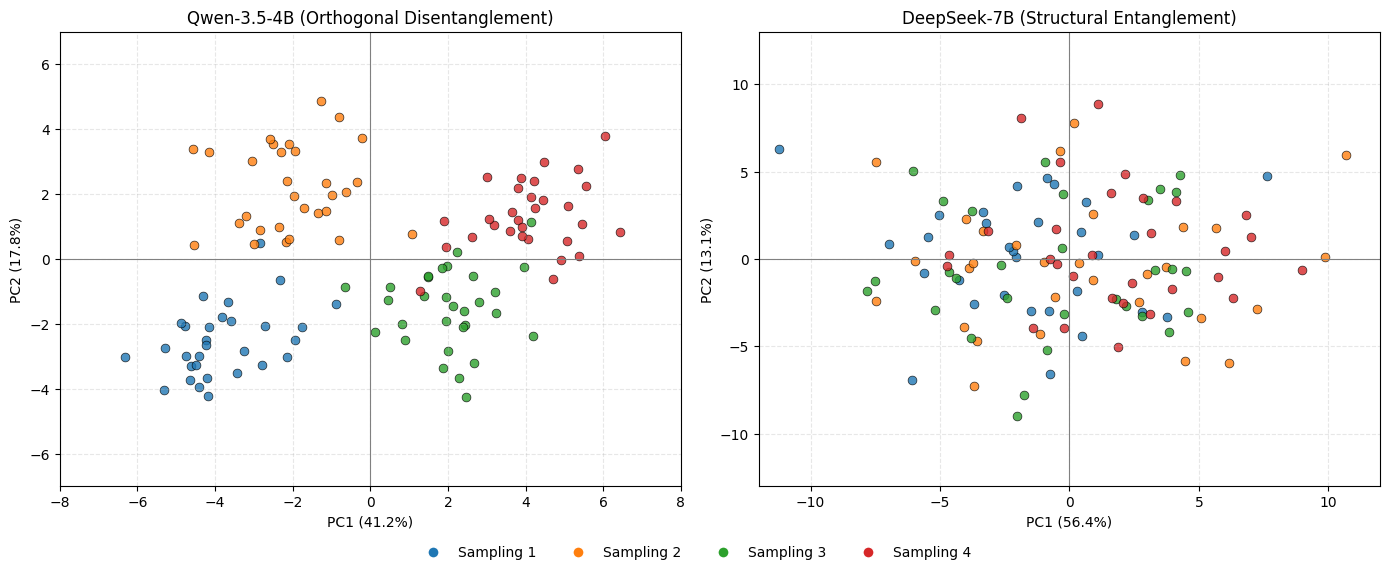}
    \caption{\textbf{PCA projections of sampled activations.}
    PCA projections of penultimate-layer activations generated via stochastic sampling. Each point represents one generated activation sample; 30 samples were collected for each of four sampling conditions (120 samples per model), with colors indicating the sampling condition. \textbf{(Left)} Qwen3.5-4B exhibits greater separation among the four sampling conditions along the leading principal components. \textbf{(Right)} DeepSeek-7B exhibits substantially greater overlap among the same conditions, consistent with the greater mixing of stylistic and structural variation that we term \textit{Structural Entanglement}.}
    \label{fig:entanglement_pca}
\end{figure}

However, \texttt{deepseek-7b-chat} presented a significant anomaly. Figure~\ref{fig:entanglement_pca} provides a complementary visualization of the activation geometry across sampling conditions. While the Qwen models successfully isolated clean stylistic axes, \texttt{deepseek-7b-chat} experienced a severe performance drop, yielding only $\sim$35\% precision. Qualitative analysis of the extracted axes for DeepSeek revealed severe entanglement between stylistic manipulation and the model's rigid structural and formatting traces.

\section{Discussion: The DeepSeek Performance Drop}
The sharp degradation of unsupervised stylistic extraction, Structural Entanglement, in \texttt{deepseek-7b-chat} highlights a fundamental architectural divergence in how certain models encode concepts. Early DeepSeek architectures, such as \texttt{deepseek-7b-chat}, were heavily optimized for strict instruction formatting, bilingual alignment, and structural compliance. When our pipeline generates the ``variation cloud'' of prompts to isolate style, \texttt{deepseek-7b-chat} exhibits substantial high-variance variation associated with these structural boundaries and output formats, rather than clearly separating stylistic variation. Consequently, structural and syntactic variance geometrically dominates the primary principal components ($PC_1$ and $PC_2$). Because our unsupervised extraction strictly isolates these top two components for semantic labeling, one hypothesis consistent with this pattern is that stylistic variance is distributed across lower-variance, unextracted singular dimensions and is therefore less recoverable by our procedure. The auto-labeling judge therefore evaluates the structural variance captured in the primary components and consequently assigns format-focused or incoherent descriptors rather than human-aligned stylistic ones, causing Precision@2 to plummet to 35.3\%. This finding suggests that certain alignment regimes natively intertwine ``how to format it'' with ``how to say it.'' This raises critical questions about the limits of zero-shot representation engineering using linear methods in architectures with heavily entrenched structural priors, suggesting that future steering methods for these models may require non-linear or layer-specific manifold learning strategies. Analogous to recent interpretability analyses demonstrating that specialized cognitive priors occupy substantial representational capacity in later architectures \citep{galichin2025icoveredbaseshere}, the structural dominance we observe is consistent with the possibility that this entanglement reflects the model's strict alignment regime rather than an artifact of our pipeline.

\section{Related Work}

\paragraph{Steering along known concepts.}
Considerable evidence suggests that high-level concepts are encoded as linear directions in LLM activation space \citep{park2024linear,tigges2024language}, though some features are irreducibly multi-dimensional \citep{engels2025not}. Activation steering exploits this structure by adding concept vectors to hidden states at inference time, deriving them from contrastive prompt pairs \citep{turner2023steering,zou2025representationengineeringtopdownapproach,rimsky2024steering}, style-labeled corpora \citep{konen2024style}, or optimization toward target completions \citep{subramani2022extracting}. Extensions condition steering on the instruction or input \citep{stolfo2025improving,lee2025programming}, compose multiple stylistic properties \citep{scalena2024multi}, and personalize outputs to user preferences \citep{cao2024personalized,bo2025steerable}, though steering reliability varies substantially across concepts and inputs \citep{tan2024analysing}; see \citet{wehner2025taxonomy} for a survey. All of these methods presuppose that the concept of interest is specified in advance, and they remain the natural choice in that setting. We study the complementary, upstream problem: identifying which stylistic dimensions are salient in a model's generations for a given prompt, without labels, curated corpora, or a predefined taxonomy. Our pipeline requires no supervision at any stage; even the polarity of each discovered axis is established automatically by the labeling judge.

\paragraph{Unsupervised feature discovery.}
Two lines of work discover features without predefined concepts. Sparse autoencoders decompose activations into large dictionaries of interpretable features \citep{cunningham2023sparse,templeton2024scaling}, but they require corpus-scale training, and recent evaluations question whether their features are canonical units or useful for downstream control \citep{wu2025axbench,kantamneni2025sparse,leask2025sparse,jorgensen2026steering}. SAE features have also been used to quantify how far representations are universal across model families \citep{lan2024quantifying}, a question our cross-model analysis revisits at the level of stylistic axes. Closer to our setting, \citet{mack2026mechanistically} discover latent behavioral modes by optimizing unsupervised perturbations to model internals, learning interpretable low-rank adapters that elicit these behaviors. Our approach differs from both: it is training-free and prompt-conditional, decomposing the model's own decoding variance via PCA to surface the few stylistic axes that dominate variation under a given instruction, rather than learning a global feature dictionary or optimizing for behavioral change.

\paragraph{Automated labeling and its validation.}
A final line of work uses LLMs to generate and score natural-language explanations of model internals at scale \citep{templeton2024scaling,paulo2024automatically}. Such labels require careful validation, as generated explanations can appear plausible without being faithful \citep{huang2023rigorously}, and LLM judges systematically favor their own generations \citep{panickssery2024llm}. The latter is a direct concern for our pipeline, where a Llama-family judge labels axes for a Llama target model; we therefore validate labels with a two-phase human study and discuss residual circularity in the Limitations section.

\section*{Conclusion}
In this work, we introduced a fully unsupervised framework for discovering human-salient stylistic axes within Large Language Models. By leveraging stochastic manifold generation and Principal Component Analysis, we demonstrated that interpretable stylistic axes can be extracted from a single prompt without any supervision. Our human evaluation shows that the discovered axes align with unconstrained human intent and that their polar generations are judged semantically valid by annotators. Overall, our results suggest that prompt-conditional stylistic structure is linearly accessible with remarkably simple tools. A natural next step is to close the loop between discovery and control: pairing unsupervised discovery with light supervised contrastive vectors \citep{rimsky2024steering} of the surfaced axes is a promising route to stylistic control without predefined taxonomies.

\section*{Limitations}

While our Unsupervised Concept Discovery pipeline demonstrates strong alignment with human stylistic intent, we acknowledge several limitations in the current framework:

\paragraph{Assumption of Linear Separability} The Latent Engine relies on Principal Component Analysis (PCA), which inherently assumes that stylistic variance and structural variance are linearly separable in the model's representation space \citep{engels2025not}. As demonstrated by the ``DeepSeek Anomaly,'' this assumption fails for architectures heavily optimized for rigid structural and syntactic compliance, resulting in structural entanglement. Extending this unsupervised discovery to models with heavily entrenched formatting priors will likely require non-linear manifold learning techniques (e.g., Kernel PCA or Sparse Autoencoders \citep{cunningham2023sparse}) to successfully disentangle features.
\paragraph{Dependence on the Linguistic Judge} The autonomous labeling phase is bottlenecked by the semantic capacity of the quantized \texttt{Meta-Llama-3-8B-Instruct} judge model. If the judge lacks the vocabulary to accurately describe a highly nuanced, purely geometric latent axis, it may assign a reductive or proximate label. While we mitigated this using an embedding synonym threshold ($\tau=0.65$), the framework remains constrained by the teacher model's linguistic ceiling and potential inherent biases. Moreover, the judge shares a model family with one evaluated target (Meta-Llama-3.2-3B). While we implemented a strictly blind evaluation protocol to prevent explicit bias, LLM evaluators can still implicitly recognize and systematically favor outputs of their own family \citep{panickssery2024llm}. Consequently, judge-assigned labels for Llama-family axes may still be slightly inflated; however, our two-phase human validation mitigates this circularity.
\paragraph{Confounding Structural Proxies} Our unsupervised extraction does not explicitly control for low-level structural covariates, such as response length, lexical diversity, or formatting markers (e.g., bullet points). Because PCA isolates the directions of maximum variance, it is possible that some discovered principal components partially align with these simpler text-surface proxies rather than purely abstract stylistic shifts.
\paragraph{Dimensional Truncation and Recall Bounds} To prioritize precision and interpretability, our pipeline isolates only the two highest-variance components ($PC_1$ and $PC_2$) from the latent manifold. However, human stylistic desire is vastly open-ended. Restricting the extraction to a strict top-$k$ truncation ($k=2$) caps the maximum possible recall. For each annotator--prompt instance $i$, the maximum recoverable fraction is $\min(2,n_i)/n_i$; averaging this quantity across the observed annotator--prompt instances yields a theoretical ceiling of approximately 51.7\% in our evaluation setup. Consequently, many valid but lower-variance stylistic features present in the model's manifold are inevitably left undiscovered by this specific configuration. Future work could explore dynamic-variance thresholding to determine an optimal, variable number of components to extract per prompt rather than utilizing a fixed cutoff.

\paragraph{Discovery versus Steering Efficacy} Our Phase 2 evaluation uses activation interventions at $\alpha=\pm0.6$ to generate the two poles of each discovered axis, and human ratings assess whether these generations express the assigned semantic trait. However, we do not evaluate steering efficacy as an independent measure of controllability—for example, whether systematically varying $\alpha$ produces monotonic changes in the target attribute or how the PCA directions compare with supervised steering directions under matched intervention strengths. In preliminary experiments, supervised contrastive vectors \citep{rimsky2024steering} produced stronger steering effects than our PCA-derived directions; our contribution is surfacing salient axes without supervision, rather than matching supervised steering strength.

\paragraph{Scope of Evaluation} Finally, our empirical evaluations were conducted exclusively on English-language generative tasks and focused strictly on stylistic, rhetorical, and affective dimensions of control. The efficacy of stochastic manifold generation for discovering localized factual knowledge, cross-lingual representations, or strict safety-alignment concepts remains an open question for future research.

\paragraph{Lower-Rank Stylistic Distribution} Because our extraction protocol strictly isolates the top two principal components ($PC_1$ and $PC_2$) to maximize clarity, we leave the empirical probing of lower-rank components to future work. Investigating whether lower-rank components conceal stylistic variance—particularly in models exhibiting Structural Entanglement like DeepSeek-7B—represents a promising direction for mapping complete model geometry.

\paragraph{Layer-Wise Selection \& Ablations} Although the penultimate layer was chosen to maintain consistent relative depth across architectures, thereby avoiding early-layer structural noise and ultimate-layer vocabulary, projection biases—our current evaluation lacks a full layer-wise ablation. Investigating how the geometric emergence of stylistic axes evolves across intermediate transformer layers remains a valuable direction for future work.

\bibliography{custom}
\clearpage
\appendix

\onecolumn
\section{Prompt Templates for Auto-Labeling}
\label{sec:appendix_prompts}

To enable the quantized Meta-Llama-3-8B-Instruct judge to autonomously label the discovered latent components, we utilized a strict, zero-shot comparative prompt. This prompt isolates the extracted geometric extremes ($-X$ and $+X$ generations) and restricts the judge to outputting a concise stylistic descriptor without extraneous reasoning. The exact template is detailed below.

\vspace{0.5em}
\noindent\fbox{%
    \parbox{\linewidth}{%
        \small
        \textbf{System Prompt:} \\
        You are an expert linguistic analyst evaluating text. Identify the SINGLE most prominent stylistic difference between Text 1 and Text 2. \\
        
        RULES: \\
        1. DO NOT name the subject matter. \\
        2. Output EXACTLY ONE WORD representing the style. \\
        3. NO conversational filler. NO sentences. \\
        
        \textbf{User Prompt:} \\
        Text 1 (Negative Pole): \\
        {[-X GENERATION]} \\
        
        Text 2 (Positive Pole): \\
        {[+X GENERATION]}
    }%
}
\vspace{0.5em}
\clearpage
\onecolumn

\section{Implementation Details}
\label{sec:appendix_implementation}

\paragraph{Hardware \& Compute Infrastructure}Due to strict compute resource constraints, all stochastic manifold generation, PCA extractions, and model evaluation pipelines were executed utilizing a Kaggle dual NVIDIA T4 GPU (2x16GB VRAM) environment. These hardware limitations necessitated our highly targeted extraction approach and the use of native 4-bit quantization for the labeling judge.

\paragraph{Software \& Libraries}
The \textit{Latent Engine} framework was implemented in Python 3.10 and relies on the following core libraries for distributed inference and geometric decomposition: PyTorch, HuggingFace Transformers, \texttt{scikit-learn} for Principal Component Analysis, and \texttt{bitsandbytes} for native 4-bit quantization.

\paragraph{Model Configuration \& Quantization}
To maximize computational throughput within the dual-T4 constraints, the pipeline utilizes an asymmetric, two-model architecture. The base target model for generation and activation extraction (\texttt{Qwen3.5-4B}) operates in native 16-bit floating-point precision (FP16). The teacher labeler model (\texttt{Meta-Llama-3-8B-Instruct}) was loaded alongside the base model using 4-bit NormalFloat (NF4) quantization.

\paragraph{Hyperparameters \& Reproducibility}
For the stochastic manifold generation phase, target architectures were sampled using an elevated temperature of $T = 0.90$ with nucleus sampling capped at $p = 0.95$. This configuration was chosen empirically to maximize the stylistic distribution of the variation cloud without deteriorating token-level syntactic coherence. The variation cloud size was strictly bounded to $N = 30$ generations per base prompt. During extraction, representations were harvested using sequence-average pooling from the penultimate layer, avoiding structural representation traps found in earlier layers. 

To account for generation stochasticity, all reported Phase 1 metrics (Section \ref{sec:aggregate_human_validation}) reflect the macro-average across 9 independent experimental seeds. The semantic alignment thresholds ($\tau = 0.65$) were computed using cosine similarity derived from the \texttt{all-mpnet-base-v2} sentence encoder.

\paragraph{Human Study Logistics and Participant Evaluation Protocol}
To ensure transparency and reproducibility, our human evaluation protocol engaged a cohort of 8 annotators with experience evaluating or producing written communication. Participants were recruited through university-based recruitment and were independent of the authors and not involved in the design or development of the Latent Engine. Participants completed both Phase 1 (unconstrained stylistic elicitation) and Phase 2 (polar validity rating on a 1--5 Likert scale) across all 20 evaluation prompts, yielding a complete evaluation matrix of 320 ordinal rating points and associated free-text semantic responses. Participants provided informed consent and were compensated for their participation.

Participants in the human evaluation study were provided with the following explicit instructions:

\begin{itemize}
    \item \textbf{Phase 1 Instructions (Spontaneous Recall):} 
    ``For each given base prompt, list up to five distinct, ideal stylistic or rhetorical control dimensions ('magic sliders') you would want available to dynamically steer the model's output generation.''
    
    \item \textbf{Phase 2 Instructions (Polar Validity):} 
    ``Evaluate the provided text generations sampled from the geometric extremes ($-X$ and $+X$) of the discovered latent axis. Rate how accurately the extreme generations embody the assigned stylistic label on a 1-5 Likert scale, where 1 indicates complete inaccuracy/incoherence and 5 indicates high accuracy.''
\end{itemize}

\paragraph{Ethics.}
The study involved evaluation of model-generated text and did not collect personally identifying or sensitive information. Participants provided informed consent prior to participation.

\clearpage
\onecolumn

\section{Evaluation Stimuli}
\label{sec:appendix_stimuli}

Table \ref{tab:prompt_matrix} details the complete evaluation matrix utilized in this study, comprising 20 distinct base prompts equally distributed across four semantic domains. 

\begin{table}[h!]
\centering
\small
\begin{tabular}{p{2.5cm} p{12.5cm}}
\toprule
\textbf{Domain} & \textbf{Base Prompt} \\
\midrule

\multirow{5}{*}{\parbox{2.5cm}{\textbf{Narrative \&\\Descriptive}}} 
& Describe the atmosphere inside a silent, ancient library using sensory details. \\
& Describe the chaotic energy of a packed night market in a bustling city. \\
& Describe a sudden thunderstorm hitting a quiet, open field during a hot afternoon. \\
& Describe a character looking out the window of a train at an unfamiliar landscape. \\
& Describe the feeling of stepping inside a cozy, warmth-filled greenhouse while it snows outside. \\
\midrule

\multirow{5}{*}{\parbox{2.5cm}{\textbf{Dialogue \&\\Creative}}} 
& Write a brief, tense exchange between two characters who are subtly trying to outsmart each other. \\
& Write a short, comforting conversation between an old mentor and a student who is ready to give up. \\
& Write a quick, witty banter session between a sarcastic adventurer and a very literal companion. \\
& Write a sharp argument between a historical scientist and a skeptical assistant regarding a new discovery. \\
& Write a mysterious dialogue snippet between a traveler and a strange gatekeeper at a mountain pass. \\
\midrule

\multirow{5}{*}{\parbox{2.5cm}{\textbf{Personal \&\\Reflexive}}} 
& Write a paragraph reflecting on the nostalgic feeling of visiting your childhood hometown. \\
& Write a brief reflection on how a sudden, quiet moment of solitude can change your perspective on a busy day. \\
& Write a short paragraph capturing the mix of anticipation and nerves before making a major life change. \\
& Write a personal paragraph about the quiet satisfaction of mastering a difficult, hands-on skill. \\
& Write a brief reflection on how an old, worn-out object can carry deep emotional meaning. \\
\midrule

\multirow{5}{*}{\parbox{2.5cm}{\textbf{Internal\\Workplace}}} 
& Write a brief, urgent Slack message asking teammates to hop onto a bridge call to fix a sudden production crash. \\
& Describe a newly proposed product feature in a short paragraph for an internal engineering documentation header. \\
& Write a concise calendar invite description for a cross-functional project sync meeting next week. \\
& Write a brief, encouraging announcement message for a team channel celebrating a successful project launch. \\
& Write a short paragraph explaining the key takeaways from a user research session to share in a leadership update chat. \\

\bottomrule
\end{tabular}
\caption{\textbf{Stimuli Matrix.} The 20 base prompts utilized for evaluation, organized by domain.}
\label{tab:prompt_matrix}
\end{table}

\clearpage

\section{Qualitative Showcase of Discovered Axes}
\label{sec:appendix_qualitative}

To demonstrate the semantic validity and geometric structure of the discovered axes, Table \ref{tab:qualitative_examples} presents unedited generation snippets sampled from the geometric extremes ($-X$ and $+X$) of the latent manifold for a representative base prompt. These examples illustrate how the PCA-derived axes map to the autonomous labels assigned by the Llama-3 judge, effectively shifting the output without requiring explicit human vocabulary.

\begin{table*}[htbp]
\centering
\small
\begin{tabular}{p{3cm} p{2cm} p{4.5cm} p{4.5cm}}
\toprule
\textbf{Base Prompt} & \textbf{Discovered Axis} & \textbf{Negative Pole ($-X$)} & \textbf{Positive Pole ($+X$)} \\
\midrule
``Write a brief, encouraging announcement message for a team channel celebrating a successful project launch.'' & \textit{Formality} & ``Hey Team! Mission accomplished with our latest success – the project has launched successfully. Thank you to everyone who contributed; your hard work really paid off. Let's celebrate this victory together and keep up that great teamwork spirit moving forward! \#TeamSpirit'' & ``Greetings esteemed colleagues,\newline\newline Our concerted efforts have culminated in triumphant success as we unveil our latest endeavor with unparalleled excellence.\newline\newline This groundbreaking triumph encapsulates not merely professional achievement but also collective spirit that has propelled us forward. The culmination of meticulous planning intertwined harmoniously with innovative execution epitomizes true teamwork's essence.\newline\newline Let this moment serve as an invigorating milestone marking the commencement of even greater heights to be achieved together within these...'' \\
\bottomrule
\end{tabular}
\caption{\textbf{Unedited Polar Generations.} Snippets extracted from the geometric extremes of the top principal components evaluated on Qwen3.5-4B. The negative pole ($\alpha = -0.6$) introduces conversational phrasing and contractions, while the positive pole ($\alpha = +0.6$) defaults to rigid corporate structures.}
\label{tab:qualitative_examples}
\end{table*}

\clearpage

\section{Single-seed Diagnostic Sensitivity Sweep}
\label{sec:appendix_threshold}

\begin{table}[h!]
\centering
\small
\begin{tabular}{lccc}
\toprule
\textbf{Threshold Criteria ($\tau$)} & \textbf{Absolute Hits ($K$)} & \textbf{Macro Precision (\%)} & \textbf{Macro Recall (\%)} \\
\midrule
$\tau = 0.40$ & 32 & 80.0\% & 44.2\% \\
$\tau = 0.50$ & 30 & 75.0\% & 41.4\% \\
$\tau = 0.60$ & 27 & 67.5\% & 37.3\% \\
\rowcolor{gray!10} \textbf{$\tau = 0.65$ (Baseline)} & \textbf{25} & \textbf{62.5\%} & \textbf{34.5\%} \\
$\tau = 0.70$ & 22 & 55.0\% & 30.4\% \\
$\tau = 0.75$ & 19 & 47.5\% & 26.2\% \\
$\tau = 0.80$ & 16 & 40.0\% & 22.1\% \\
$\tau = 0.85$ & 14 & 35.0\% & 19.3\% \\
$\tau = 0.90$ & 12 & 30.0\% & 16.6\% \\
$\tau = 0.95$ & 11 & 27.5\% & 15.2\% \\
\midrule
$\tau = 1.00$ (Exact Match) & 10 & 25.0\% & 13.8\% \\
\bottomrule
\end{tabular}
\caption{\textbf{Sensitivity sweep over the semantic matching parameter $\tau$.} The macro precision maps strictly to the discrete hits within a single canonical evaluation dimension cohort ($D=40$). While our top-performing architecture (Qwen3.5-4B) achieves a higher aggregate baseline of 72.8\% macro-averaged across 9 seeds, this specific sample run (62.5\% at $\tau=0.65$) isolates a single seed to profile the smooth mathematical decay of the alignment framework without multi-seed averaging artifacts.}
\label{tab:tau_sensitivity}
\end{table}

\clearpage

\section{Extended Evaluation Metrics}
\label{sec:appendix_eval}

\begin{figure}[h!]
    \centering
    \includegraphics[width=\textwidth]{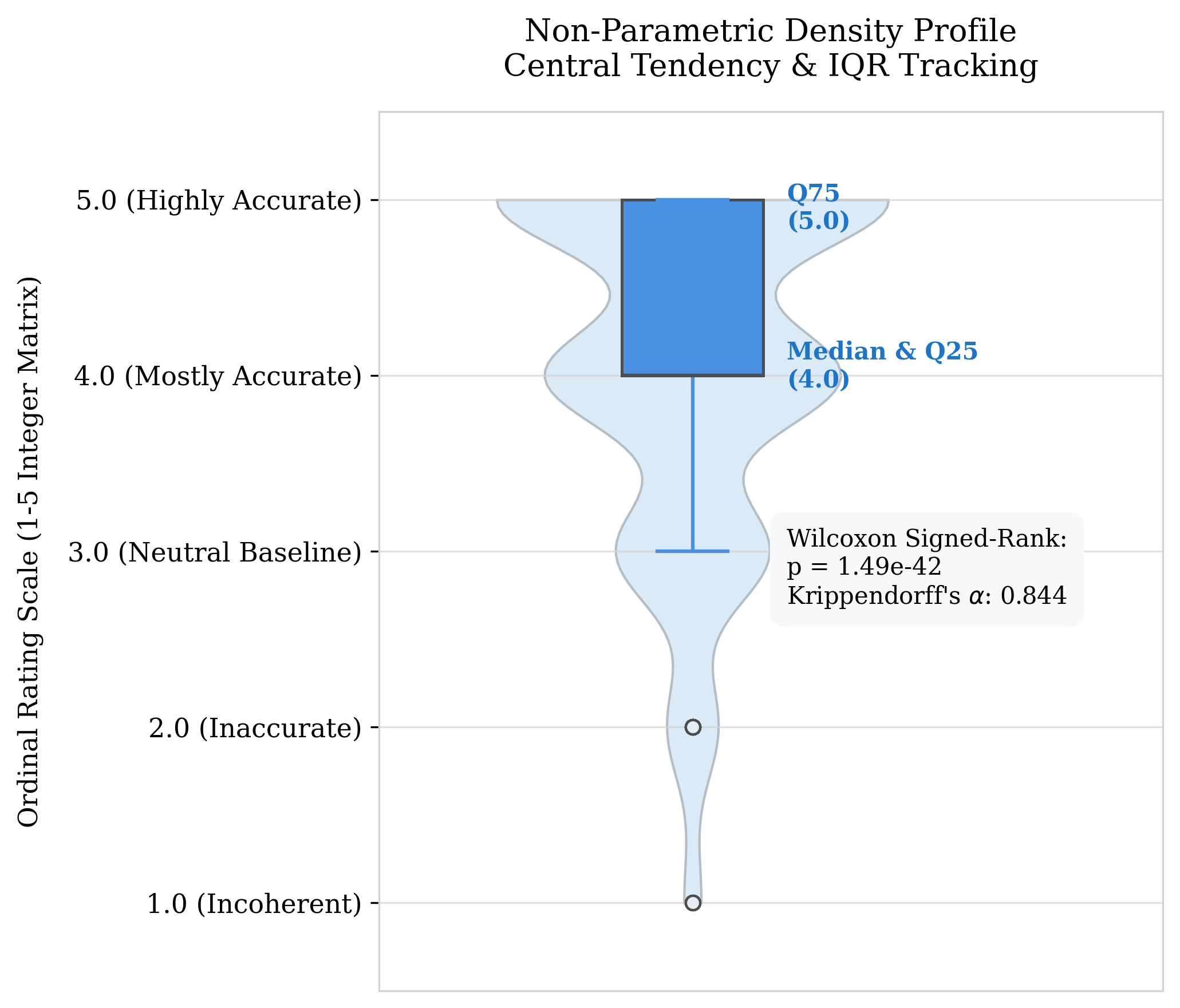}
    \caption{\textbf{Non-Parametric Density Profile of Phase 2 Human Evaluations.} A violin plot with an embedded box plot illustrating the distribution of ordinal validity ratings ($N=320$) for the auto-discovered stylistic axes. Because Likert data is strictly ordinal, this non-parametric visualization captures the true central tendency without assuming a normal distribution. The density profile reveals a severe left-skew concentrated at the upper bounds of the scale: both the median and the first quartile ($Q_{25}$) are anchored at 4.0 (``Mostly Accurate''), with the third quartile ($Q_{75}$) at 5.0 (``Highly Accurate''). The overlaid statistical markers confirm the robustness of this alignment: a Wilcoxon Signed-Rank test ($p = 1.49 \times 10^{-42}$) rejects the null hypothesis of neutral or random rating assignment, while Krippendorff's $\alpha = 0.844$ indicates a strong inter-annotator agreement. This provides granular structural support for the global top-2 box accuracy reported in Section 3.1.}
    \label{fig:violin_density}
\end{figure}

\end{document}